\documentclass{article}

\usepackage{microtype}
\usepackage{graphicx}
\usepackage{subcaption}
\usepackage{booktabs} 

\usepackage{hyperref}

\usepackage[preprint]{icml2026}

\usepackage{amsmath}
\usepackage{amssymb}
\usepackage{mathtools}
\usepackage{amsthm}
\usepackage{adjustbox}
\usepackage{wrapfig}
\usepackage{pifont}

\graphicspath{{./figs/}}
\usepackage{colortbl} 

\usepackage{tikz}
\usepackage{pgfplots}
\pgfplotsset{compat=1.17}

\usepackage{url}

\usepackage{tikz}

\usepackage{multirow}
\newcommand{\minisection}[1]{\vspace{.04in}\noindent{\textbf{#1}.}}

\usepackage[capitalize,noabbrev]{cleveref}

\usepackage{tabularray}
\usepackage{multirow}
\usepackage{array, booktabs}
\newcolumntype{C}[1]{>{\centering\let\newline\\\arraybackslash\hspace{0pt}}m{#1}}
\newcolumntype{M}[1]{>{\centering\arraybackslash}m{#1}}
\usepackage{anyfontsize}

\usepackage{color, colortbl}
\definecolor{Gray}{gray}{0.9}
\definecolor{DarkGray}{rgb}{0.75, 0.75, 0.75}
\definecolor{LightCyan}{gray}{0.95}
\newcolumntype{g}{>{\columncolor{Gray}}c}

\renewcommand{\algorithmicrequire}{\textbf{Input:} }   
\renewcommand{\algorithmicensure} {\textbf{Output:}}

\theoremstyle{plain}

\theoremstyle{definition}

\theoremstyle{remark}

\usepackage[textsize=tiny]{todonotes}

\icmltitlerunning{From Pruning to Grafting: Dynamic Knowledge Redistribution via Learnable Layer Fusion}

\begin{document}

\twocolumn[
  \icmltitle{From Pruning to Grafting: Dynamic Knowledge Redistribution via Learnable Layer Fusion}



  \icmlsetsymbol{equal}{*}

\begin{icmlauthorlist}
\icmlauthor{Zehua Pei}{cuhk,noah}
\icmlauthor{Hui-Ling Zhen}{noah}
\icmlauthor{Xianzhi Yu}{noah}
\icmlauthor{Sinno Jialin Pan}{cuhk}
\icmlauthor{Mingxuan Yuan}{noah}
\icmlauthor{Bei Yu}{cuhk}
\end{icmlauthorlist}

\icmlaffiliation{cuhk}{The Chinese University of Hong Kong, Hong Kong SAR}
\icmlaffiliation{noah}{Noah’s Ark Lab, Huawei, Hong Kong SAR}

\icmlcorrespondingauthor{Zehua Pei}{zhpei23@cse.cuhk.edu.hk}

  \icmlkeywords{Machine Learning, ICML}

  \vskip 0.3in
]



\printAffiliationsAndNotice{}  

\begin{abstract}
Structured pruning of Generative Pre-trained Transformers (GPTs) offers a promising path to efficiency but often suffers from irreversible performance degradation due to the discarding of transformer blocks. In this paper, we introduce FuseGPT, a compression paradigm that reframes structured pruning as iterative knowledge grafting rather than simple removal. Motivated by the observation that linear block merging fails to capture non-linear feature disparities and that block importance fluctuates dynamically during pruning, FuseGPT employs a dual-strategy pipeline. First, we propose Macro Influence (MI), a dynamic fusion-aware metric that continuously re-evaluates block redundancy as the network topology evolves. Second, instead of rigid parameter averaging, we introduce a learnable low-rank fusion mechanism that adaptively grafts the knowledge of pruned blocks onto surviving layers via lightweight local distillation. Extensive experiments on LLaMA, Mistral, Qwen, and Phi families demonstrate that FuseGPT establishes a new state-of-the-art on the compression-accuracy Pareto frontier: at 25\% sparsity, FuseGPT achieves lower perplexity than prior methods at 20\% sparsity, improves zero-shot reasoning by up to 4.5 points, and delivers 1.33$\times$ inference speedup with 25\% memory reduction. Furthermore, FuseGPT is orthogonal to quantization, achieving 52.1\% total compression with negligible quality loss when combined with 4-bit GPTQ.
We make our code publicly available at \url{https://github.com/JarvisPei/FuseGPT}.
\end{abstract}


\section{Introduction}
\label{sec:intro}

Generative Pre-trained Transformers (GPTs) have demonstrated remarkable capabilities in handling complex tasks and exhibiting emergent abilities in various domains, especially when scaled to billions of parameters~\cite{brown2020language, zhang2022opt, touvron2023llama, liu2024visual}.
Despite their unprecedented success, the increasing complexity and size of GPTs have introduced significant challenges for deployment in real-world scenarios, particularly in resource-constrained environments.

To address the hardware demands associated with deploying GPTs, model compression techniques are developed to produce more compact models while preserving high performance.
These techniques primarily fall into two categories: model pruning and quantization~\cite{lecun1989optimal, han2015learning, hoefler2021sparsity, liu2021post, gholami2022survey}.
This paper focuses on model pruning, a technique aimed at reducing model size by eliminating redundant parameters.
Pruning is mainly categorized into two types: unstructured pruning and structured pruning~\cite{frantar2023sparsegpt, wang2019structured}. 
Unstructured pruning targets removing individual weights, which can achieve higher performance but often results in hardware-unfriendly sparse weights, limiting acceleration potential.
Structured pruning, in contrast, removes entire pre-defined structures (e.g., layers or blocks) at once, which may lead to a slight reduction in accuracy but is more hardware-efficient.

Recent studies have revealed that redundancy exists across transformer blocks in GPTs,
meaning that certain blocks contribute less significantly to the final outcomes~\cite{men2024shortgpt, kim2024shortened, song2024sleb}.
Some existing methods detect this redundancy by analyzing the similarities between hidden states, while others directly measure the changes in distance to the hard labels.
Once redundant blocks are identified, structured pruning is applied to remove the least important ones, aiming to minimize the performance degradation.
However, simply discarding these blocks often results in irreversible performance loss.
While traditional post-pruning fine-tuning can help recover the performance, they typically require extensive datasets and substantial computational resources. As a result, there is a pressing need for more efficient methods to restore model performance without such heavy demands.

\begin{figure*}[tb!]
    \centering
    \includegraphics[width=0.82\linewidth]{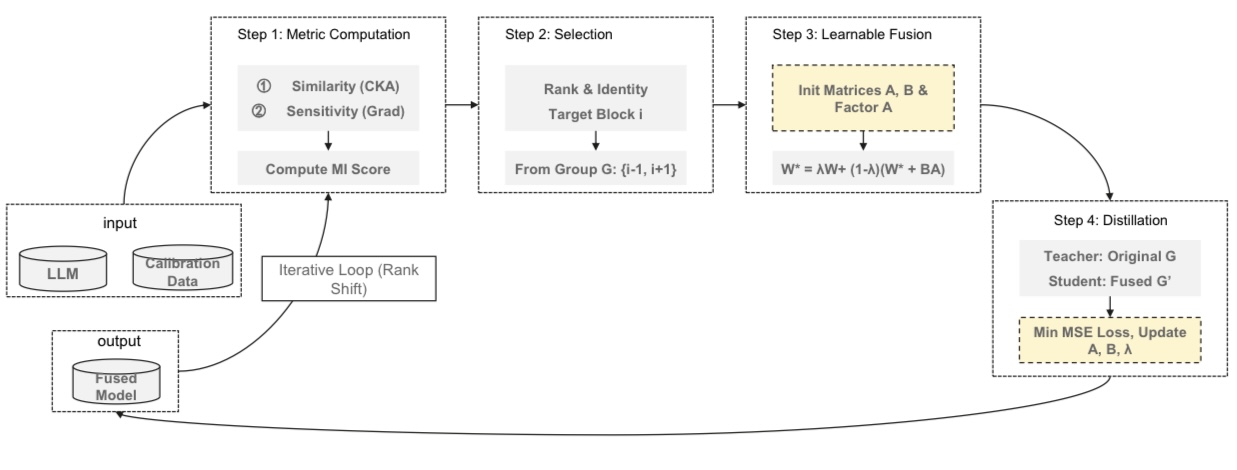} 
    \caption{The overall pipeline of the FuseGPT. The process is iterative to address rank shift. (1) Metric: We compute Macro Influence combining similarity and sensitivity. (2) Selection: The least influential block is selected and grouped with neighbors. (3) Fusion: Knowledge is grafted using learnable low-rank matrices (A, B) and interpolation factor $\lambda$. (4) Distillation: Only the local group is fine-tuned using MSE loss to align student and teacher representations.}
    \label{fig:fusegpt}
\end{figure*}

In this paper, we introduce \textbf{FuseGPT}, a novel prune-and-fuse compression paradigm designed to overcome the limitations of static pruning. As illustrated in Figure~\ref{fig:fusegpt}, unlike one-shot methods that overlook the evolving nature of layer importance, our framework explicitly addresses the \textit{Rank Shift} phenomenon---where the relative criticality of remaining layers fluctuates after a block is removed. To handle this, FuseGPT operates in an iterative four-step cycle: (1) \textbf{Metric Calculation:} We first evaluate the \textit{Macro Influence} of each block, a composite metric that synthesizes layer similarity and sensitivity to capture both representational redundancy and functional importance. (2) \textbf{Dynamic Selection:} Based on these updated scores, we identify the least critical block and its immediate neighbors, acknowledging that layer importance shifts dynamically after every modification. (3) \textbf{Learnable Fusion:} Instead of hard removal, we employ a lightweight fusion mechanism. The target block is \textit{grafted} onto its neighbor via low-rank adaptation, preserving essential knowledge that would otherwise be discarded. (4) \textbf{Distillation and Local Healing:} Finally, we perform a rapid, localized distillation step to align the fused block's output with the original teacher model, ensuring seamless integration before the next iteration begins.
This cycle repeats until the target compression ratio is met, ensuring robustness even at high pruning rates.

Our key contributions are summarized as follows:

\begin{itemize}
    \item We introduce \textbf{Macro Influence (MI)}, a dynamic importance metric designed to identify blocks best suited for \textbf{absorption} by their neighbors. Unlike static heuristics, MI is updated iteratively to counter the ``rank shift'' effect, ensuring that detection and recovery remain synergistic throughout the compression process.

    \item We propose a \textbf{learnable low-rank fusion mechanism} that \textbf{recycles} knowledge from pruned blocks. By grafting parameters onto surviving neighbors via adaptive coefficients and guiding them with partial-group distillation, we avoid the feature collapse common in rigid averaging or interpolation schemes.
    
    \item We demonstrate that FuseGPT establishes a new state-of-the-art on the \textbf{compression-accuracy Pareto frontier}. It significantly outperforms prior layer-merging methods (e.g., LaCo, MKA) across diverse architectures (LLaMA-3, Mistral, Qwen) and multimodal models (LLaVA-1.5), achieving up to a $27\%$ reduction in perplexity compared to baselines.
    
    \item We show that FuseGPT is \textbf{orthogonal to quantization and evolutionary search}. It achieves $52.1\%$ total compression when combined with 4-bit GPTQ and further extends the efficiency frontier when integrated with evolutionary algorithms (FuseGPT-Evo), paving the way for extreme model compression.
\end{itemize}

Extensive experiments confirm that FuseGPT delivers superior performance across the board. Beyond achieving a \textbf{33\% inference speedup} and substantial perplexity reduction, our method exhibits remarkable robustness in zero-shot evaluations across both language and multimodal tasks, proving its viability for efficient, high-performance deployment.

\section{Related Work}

To reduce the inference cost of large language models (LLMs), numerous studies have focused on model compression, primarily through pruning and quantization~\cite{xia2022structured, kurtic2022optimal, ma2023llm, yao2022zeroquant, lin2024awq, zou2024bie, louizos2017learning}. We review the most relevant lines of work below.

\minisection{Structured Pruning}
Pruning methods identify and remove redundant parameters to obtain sparser models. Unstructured pruning~\cite{frantar2023sparsegpt} achieves high compression but yields irregular sparsity patterns that are difficult to accelerate on commodity hardware. Structured pruning, in contrast, removes entire architectural units. ShortGPT~\cite{men2024shortgpt} proposes Block Influence (BI), measuring layer redundancy via input-output similarity, and removes low-BI layers directly. SLEB~\cite{song2024sleb} refines this by evaluating block importance through end-to-end prediction loss. SliceGPT~\cite{ashkboos2024slicegpt} takes an orthogonal approach, reducing the embedding dimension by projecting weight matrices into lower-dimensional subspaces. FoldGPT~\cite{foldgpt} combines layer removal with parameter sharing across retained blocks. These methods typically treat pruning as \textit{removal}: once a block is identified as redundant, its parameters are discarded. FuseGPT departs from this paradigm by treating pruning as \textit{redistribution}, recycling the knowledge of pruned blocks into their neighbors.

\minisection{Layer Merging}
Recent work has explored merging multiple transformer layers into fewer blocks as an alternative to outright removal. LaCo~\cite{yang2024laco} collapses adjacent layers based on representation similarity metrics (e.g., RDSC), using linear interpolation to combine parameters. MKA~\cite{liu2024pruning} identifies mergeable layers through manifold alignment in activation space and applies closed-form averaging. While effective, these approaches rely on static, predetermined fusion rules. FuseGPT extends this line of work by introducing \textit{learnable} fusion: instead of fixed interpolation coefficients, we optimize low-rank adaptation matrices that graft pruned knowledge onto surviving blocks, guided by a fusion-aware importance metric.

\minisection{Knowledge Distillation}
Knowledge distillation (KD) transfers knowledge from a large teacher model to a smaller student~\cite{hinton2015distilling}. In the context of language models, DistilBERT~\cite{sanh2019distilbert} and TinyBERT~\cite{jiao2020tinybert} distill BERT into compact students by matching hidden states and attention distributions. Recent LLM distillation efforts~\cite{gu2024minillm, agarwal2024onpolicy} focus on aligning output distributions or chain-of-thought reasoning.

\section{Motivation}
~\label{sec:motivation}

Before detailing our methodology, we present two pilot studies that challenge the core assumptions of existing pruning and fusion techniques. These observations motivate the design of FuseGPT.

\textbf{Observation I: High Activation Similarity $\neq$ Successful Linear Fusion.}
Prior methods like LaCo~\cite{yang2024laco} operate on an implicit \textit{Linearity Assumption}: if two blocks $B_i$ and $B_j$ exhibit high cosine similarity in their output activations, their weight matrices can be merged via linear interpolation (e.g., $\alpha W_i + (1-\alpha)W_j$) without significant information loss.

To systematically test this assumption, we conducted an experiment on LLaMA-2-7B. For all pairs of adjacent blocks $(B_i, B_{i+1})$ where $i \in \{1, \ldots, 31\}$, we computed: (1) the cosine similarity between their output activations on a calibration set, and (2) the mean squared error (MSE) between the output of a linearly merged block and the original two-block output. If the linearity assumption holds, high activation similarity should correlate with low fusion MSE. 
As shown in Figure~\ref{fig:observation}(a), this correlation is surprisingly weak (Pearson $r = 0.23$). Many block pairs with activation similarity exceeding 0.85 still exhibit high fusion MSE ($>0.3$), resulting in perplexity spikes when merged. We attribute this to \textbf{Destructive Interference}: even when two blocks produce similar activations, their weight matrices may lie on different functional manifolds. Linear averaging disrupts the precise transformations learned by each block, leading to feature collapse. This suggests that fusion should be treated as a \textit{learned adaptation} problem rather than a closed-form averaging operation.

\begin{figure}[htbp]
    \centering
    \begin{subfigure}[t]{0.48\linewidth}
        \centering
        \resizebox{\linewidth}{!}{
        \begin{tikzpicture}
            \begin{axis}[
                xlabel={Activation Similarity},
                ylabel={Fusion MSE},
                xmin=0.5, xmax=1.0,
                ymin=0, ymax=0.6,
                grid=major,
                grid style={dashed, gray!30},
                xlabel style={font=\small},
                ylabel style={font=\small},
                tick label style={font=\footnotesize},
                width=6.5cm,
                height=5.5cm,
            ]
            \addplot[only marks, mark=*, mark size=2pt, blue!70, opacity=0.8] coordinates {
                (0.52, 0.48) (0.55, 0.51) (0.58, 0.44) (0.61, 0.39)
                (0.63, 0.42) (0.65, 0.36) (0.67, 0.38) (0.69, 0.33)
            };
            \addplot[only marks, mark=*, mark size=2pt, blue!70, opacity=0.8] coordinates {
                (0.71, 0.35) (0.73, 0.29) (0.74, 0.32) (0.76, 0.28)
                (0.77, 0.31) (0.78, 0.25) (0.79, 0.33) (0.80, 0.22)
                (0.81, 0.27) (0.82, 0.30) (0.83, 0.19) (0.84, 0.26)
            };
            \addplot[only marks, mark=*, mark size=2.5pt, red!70, opacity=0.9] coordinates {
                (0.86, 0.34) (0.87, 0.31) (0.88, 0.38) (0.89, 0.29)
                (0.90, 0.42) (0.91, 0.25) (0.92, 0.33) (0.93, 0.21)
            };
            \addplot[only marks, mark=*, mark size=2pt, blue!70, opacity=0.8] coordinates {
                (0.85, 0.18) (0.88, 0.14) (0.91, 0.11) (0.94, 0.08)
                (0.95, 0.12) (0.96, 0.09) (0.97, 0.06) (0.98, 0.05)
            };
            \addplot[dashed, thick, black!60, domain=0.5:1.0] {0.55 - 0.35*x};
            \node[font=\footnotesize, align=center] at (axis cs:0.89, 0.52) {High similarity,\\high MSE};
            \draw[->, thick, gray] (axis cs:0.89, 0.48) -- (axis cs:0.90, 0.42);
            \node[font=\footnotesize, fill=white, inner sep=2pt] at (axis cs:0.65, 0.08) {$r = 0.23$};
            \end{axis}
        \end{tikzpicture}
        }
        \caption{Activation similarity vs. fusion MSE for all adjacent block pairs in LLaMA-2-7B. Red points highlight pairs with high similarity ($>0.85$) but high fusion error, violating the linearity assumption.}
        \label{fig:obs-scatter}
    \end{subfigure}\hfill
    \begin{subfigure}[t]{0.48\linewidth}
        \centering
        \resizebox{\linewidth}{!}{
        \begin{tikzpicture}
            \begin{axis}[
                xlabel={Rank Before Removal},
                ylabel={Rank After Removal},
                xmin=0, xmax=32,
                ymin=0, ymax=32,
                grid=major,
                grid style={dashed, gray!30},
                xlabel style={font=\small},
                ylabel style={font=\small},
                tick label style={font=\footnotesize},
                width=6.5cm,
                height=5.5cm,
                xtick={0,8,16,24,32},
                ytick={0,8,16,24,32},
            ]
            \addplot[thick, black!40, dashed, domain=1:31] {x};
            \addplot[only marks, mark=*, mark size=2.5pt, blue!80] coordinates {
                (1, 3) (2, 1) (3, 5) (4, 2) (5, 9) (6, 4) (7, 11) (8, 6)
                (9, 7) (10, 15) (11, 8) (12, 18) (13, 10) (14, 12) (15, 21)
                (16, 13) (17, 14) (18, 23) (19, 16) (20, 17) (21, 25) (22, 19)
                (23, 20) (24, 8) (25, 22) (26, 27) (27, 24) (28, 26) (29, 28)
                (30, 29) (31, 30)
            };
            \addplot[only marks, mark=*, mark size=3.5pt, red!80] coordinates {
                (24, 8) (5, 9) (10, 15) (12, 18) (15, 21)
            };
            \draw[->, thick, red!60] (axis cs:24, 12) -- (axis cs:24, 8.5);
            \node[font=\tiny, red!80, align=center] at (axis cs:27, 11) {Rank 24\\$\rightarrow$ 8};
            \node[font=\footnotesize, fill=white, inner sep=2pt] at (axis cs:8, 28) {$\rho = 0.58$};
            \node[font=\tiny, black!50, rotate=45] at (axis cs:28, 30) {$\rho = 1.0$};
            \end{axis}
        \end{tikzpicture}
        }
        \caption{Block importance ranking before vs. after removing one block. Deviation from the diagonal indicates rank shift. Red points show blocks with $>10$ rank change.}
        \label{fig:obs-rank}
    \end{subfigure}
    \caption{Pilot studies on LLaMA-2-7B challenging assumptions of existing methods. (a) Weak correlation between activation similarity and fusion success suggests linear merging is unreliable. (b) Significant rank shift after single block removal demonstrates that importance scores are topology-dependent, invalidating one-shot pruning strategies.}
    \label{fig:observation}
\end{figure}

\textbf{Observation II: Block Importance is Dynamic, Not Static.}
Existing structured pruning methods~\cite{men2024shortgpt, ashkboos2024slicegpt, song2024sleb} typically employ a \textit{one-shot ranking} strategy: compute importance scores for all blocks once, then prune the lowest-ranked blocks simultaneously. This approach implicitly assumes that block importance is an intrinsic property, independent of the network's current topology.

We challenge this assumption with the following experiment on LLaMA-2-7B. We first compute importance scores (using the Macro Influence metric defined in Section~\ref{sec:method}) for all 32 blocks and record their rankings. We then remove the single least important block and immediately recompute importance scores for the remaining 31 blocks. If importance were static, the relative ranking of surviving blocks should be preserved.
Figure~\ref{fig:observation}(b) reveals substantial \textit{rank shift}: the Spearman correlation between pre- and post-removal rankings is only $\rho = 0.58$. Blocks previously deemed redundant (e.g., ranked 25th) jumped to critical status (e.g., ranked 8th) after their neighbor was removed, while some previously important blocks became more dispensable. This demonstrates that block importance is highly contextual—removing one block redistributes computational burden across the network, fundamentally altering the importance landscape. Relying on stale, pre-computed scores leads to suboptimal pruning decisions, particularly when multiple blocks are removed. This observation necessitates an \textit{iterative} compression strategy that re-evaluates importance after each modification.

\section{Methodology} 
\label{sec:method}

Based on the observations above, we formally introduce \textbf{FuseGPT}, a compression paradigm that reframes structural pruning as a knowledge redistribution problem. Unlike traditional methods that discard redundant blocks, FuseGPT iteratively identifies blocks with high \textit{replaceability} and grafts their functional knowledge onto neighboring blocks via learnable manifold alignment.

\subsection{Problem Formulation}

Consider a pre-trained Transformer model $\mathcal{M}$ consisting of $N$ stacked blocks $\{B_1, \dots, B_N\}$, where each block $B_i$ transforms hidden states $\mathbf{H}_{i-1} \in \mathbb{R}^{L \times d}$ to $\mathbf{H}_i \in \mathbb{R}^{L \times d}$, with $L$ denoting the sequence length and $d$ the hidden dimension. Our goal is to obtain a compressed model $\mathcal{M}'$ with $N' < N$ blocks such that the divergence between the predictive distributions of $\mathcal{M}$ and $\mathcal{M}'$ is minimized.

We formulate this as an iterative optimization problem. At each step $t$, we seek to identify a block $B_p$ and a set of optimal fusion parameters $\Theta$ such that removing $B_p$ while updating its neighbors minimizes the information loss:
\begin{equation}
\label{eq:objective}
    \min_{p, \Theta} \mathcal{L}_{\text{div}}\left(\mathcal{M}(\mathbf{X}), \textsc{Fuse}(\mathcal{M}_{\setminus p}, \Theta; \mathbf{X})\right),
\end{equation}
where $\mathbf{X} \in \mathbb{R}^{L \times d}$ denotes the input hidden states, $\mathcal{M}_{\setminus p}$ denotes the model with block $B_p$ removed, and $\textsc{Fuse}(\cdot)$ represents our learnable fusion operator that redistributes the knowledge of $B_p$ to its neighboring blocks.

\subsection{Fusion-Aware Importance Scoring}
\label{sec:mi}

Standard importance metrics (e.g., weight magnitude, Taylor expansion) typically measure sensitivity to removal in isolation. However, for a prune-and-fuse paradigm, the critical criterion is not just redundancy, but \textit{absorbability}—how easily a block's function can be compensated by its neighbors. We propose \textbf{Macro Influence (MI)}, a fusion-aware metric that evaluates the global impact of block removal on the final representation manifold.

For each block $B_i$, we construct a temporary model $\mathcal{M}_{\setminus i}$ by removing $B_i$ and directly connecting $B_{i-1}$ to $B_{i+1}$. We then measure the perturbation in the final hidden states caused by this removal. Specifically, let $\mathbf{H}^{(N)}_{\mathcal{M}}$ denote the output of the last block in the original model $\mathcal{M}$, and $\mathbf{H}^{(N-1)}_{\mathcal{M}_{\setminus i}}$ denote the corresponding output from the pruned model $\mathcal{M}_{\setminus i}$. We define the Macro Influence of block $B_i$ as:
\begin{equation}
\label{eq:mi}
    \text{MI}(B_i) = 1 - \mathbb{E}_{\mathbf{x} \sim \mathcal{D}_{\text{cal}}} \left[ \cos\left(\mathbf{H}^{(N)}_{\mathcal{M}}(\mathbf{x}),\, \mathbf{H}^{(N-1)}_{\mathcal{M}_{\setminus i}}(\mathbf{x})\right) \right],
\end{equation}
where $\mathcal{D}_{\text{cal}}$ is a small calibration dataset, and $\cos(\cdot, \cdot)$ computes the cosine similarity averaged over all token positions.

A lower $\text{MI}(B_i)$ indicates that removing $B_i$ causes minimal perturbation to the final representations, suggesting that its semantic contribution is either redundant or can be implicitly reconstructed by the remaining network. Such blocks are ideal candidates for fusion. Crucially, unlike metrics based on hard-label loss, MI leverages the dense information in soft hidden states, which has been shown to better preserve generalization capability during compression~\citep{hinton2015distilling}.

\subsection{Learnable Manifold Grafting}
\label{sec:fusion}

Once the candidate block $B_p = \arg\min_{i} \text{MI}(B_i)$ is identified, we proceed to redistribute its knowledge to neighboring blocks. As demonstrated in Section~\ref{sec:motivation}, direct linear averaging of weights leads to feature collapse because weight matrices of distinct blocks lie on different functional manifolds. To address this, we propose \textbf{Learnable Manifold Grafting}, which injects the pruned block's knowledge into its neighbors via learnable transformations.

\paragraph{Fusion Window.} We define a local fusion window $\mathcal{B}_{\text{local}}$ of size $G$ centered around the pruned block $B_p$:
\begin{equation}
\label{eq:partial}
    \mathcal{B}_{\text{local}} = \{B_{p-\lfloor G/2 \rfloor}, \dots, B_{p-1}, B_{p+1}, \dots, B_{p+\lceil G/2 \rceil}\} \setminus \{B_p\},
\end{equation}
where boundary conditions are handled by truncation. The fusion window determines which blocks will absorb the knowledge from $B_p$.

\paragraph{Coefficient-Based Knowledge Injection.} For each linear layer in $B_p$ with weight matrix $\mathbf{W}_p \in \mathbb{R}^{d_{\text{out}} \times d_{\text{in}}}$, we inject its information into the corresponding layer of each neighbor block $B_i \in \mathcal{B}_{\text{local}}$ via a learnable coefficient matrix. Specifically, the fused weight matrix is computed as:
\begin{equation}
\label{eq:grafting}
    \mathbf{W}^{\text{fused}}_{i} = \mathbf{W}_{i} + \mathbf{C} \odot \mathbf{W}_{p},
\end{equation}
where $\mathbf{W}_{i} \in \mathbb{R}^{d_{\text{out}} \times d_{\text{in}}}$ is the original weight matrix of the neighbor block, $\mathbf{C} \in \mathbb{R}^{d_{\text{out}} \times d_{\text{in}}}$ is a learnable coefficient matrix, and $\odot$ denotes the Hadamard (element-wise) product. This formulation allows selective incorporation of features from $B_p$: entries of $\mathbf{C}$ close to zero preserve the original behavior of $B_i$, while non-zero entries graft specific transformations from $B_p$.

\paragraph{Low-Rank Parameterization.} To ensure parameter efficiency and prevent overfitting, we impose a low-rank constraint on the coefficient matrix by decomposing it as:
\begin{equation}
\label{eq:lowrank}
    \mathbf{C} = \mathbf{A}\mathbf{B}^\top,
\end{equation}
where $\mathbf{A} \in \mathbb{R}^{d_{\text{out}} \times r}$ and $\mathbf{B} \in \mathbb{R}^{d_{\text{in}} \times r}$ are learnable low-rank factors, and $r \ll \min(d_{\text{out}}, d_{\text{in}})$ is the rank hyperparameter. The resulting coefficient matrix $\mathbf{C} = \mathbf{A}\mathbf{B}^\top \in \mathbb{R}^{d_{\text{out}} \times d_{\text{in}}}$ has the same shape as $\mathbf{W}_p$, ensuring compatibility for the Hadamard product in Eq.~\eqref{eq:grafting}.

Substituting Eq.~\eqref{eq:lowrank} into Eq.~\eqref{eq:grafting}, the forward pass of the fused layer becomes:
\begin{equation}
\label{eq:forward}
    \mathbf{Y} = \mathbf{W}^{\text{fused}}_{i} \mathbf{X} = \mathbf{W}_{i}\mathbf{X} + \left(\mathbf{A}\mathbf{B}^\top\right) \odot \left(\mathbf{W}_{p}\mathbf{X}\right),
\end{equation}
where $\mathbf{X} \in \mathbb{R}^{d_{\text{in}} \times L}$ is the input activation and $\mathbf{Y} \in \mathbb{R}^{d_{\text{out}} \times L}$ is the output.

\paragraph{Initialization Strategy.} We initialize $\mathbf{A}$ with zeros and $\mathbf{B}$ with Kaiming uniform initialization~\citep{he2015delving}. This ensures that $\mathbf{C} = \mathbf{A}\mathbf{B}^\top = \mathbf{0}$ at the start of optimization, so the fused layer initially behaves identically to the original layer of $B_i$. The optimization then gradually introduces knowledge from $B_p$ as needed.

\paragraph{Recursive Fusion.} In iterative pruning, a neighbor block $B_i$ may have already absorbed knowledge from previously pruned blocks. To handle this, we maintain the accumulated weight state: before grafting $B_p$'s knowledge, we first compute the effective weights of any previously fused block and treat them as frozen during the current fusion step.

\subsection{Local Distillation}
\label{sec:distill}

To optimize the low-rank factors $\{\mathbf{A}, \mathbf{B}\}$ and adapt the survivor blocks, we employ a lightweight, localized distillation process. Rather than fine-tuning the entire model, we isolate the fusion window and minimize the divergence between the original block sequence's output and the fused blocks' output.
Let $\mathbf{H}_{\text{orig}}$ denote the output hidden states when the input passes through the original block sequence $\mathcal{B}_{\text{local}} \cup \{B_p\}$, and let $\mathbf{H}_{\text{fused}}$ denote the output when passing through the fused blocks $\mathcal{B}^{\text{fused}}_{\text{local}}$ (with $B_p$ removed). We minimize the following KL divergence loss:
\begin{equation}
\label{eq:distill}
    \mathcal{L}_{\text{distill}} = \text{KL}\left( \sigma(\mathbf{H}_{\text{orig}} / \tau) \,\|\, \sigma(\mathbf{H}_{\text{fused}} / \tau) \right),
\end{equation}
where $\sigma(\cdot)$ denotes the softmax function applied along the feature dimension, and $\tau$ is a temperature hyperparameter that controls the smoothness of the distribution. This local adaptation ensures that the fused blocks functionally mimic the original sequence, effectively compensating for the depth reduction.

\paragraph{Weight Folding.} After distillation converges, we fold the learned coefficients into the weight matrices for zero-overhead inference:
\begin{equation}
\label{eq:fold}
    \mathbf{W}_{i} \leftarrow \mathbf{W}_{i} + \left(\mathbf{A}\mathbf{B}^\top\right) \odot \mathbf{W}_{p}.
\end{equation}
The low-rank factors $\mathbf{A}, \mathbf{B}$ and the pruned block $B_p$ are then discarded, yielding a compressed model with no additional parameters or computational overhead at inference time.

Algorithm~\ref{algo:fusegpt} summarizes the complete FuseGPT pipeline. Given a target sparsity ratio $S$, we iteratively: (1) compute MI scores for all remaining blocks, (2) identify the block with lowest MI as the fusion candidate, (3) graft its knowledge onto neighbors via learnable manifold grafting, and (4) remove the candidate block. This process repeats until the desired compression ratio is achieved. The subroutine for group fusion is detailed in Algorithm~\ref{algo:subroutine}.

\begin{algorithm}[t]
\caption{FuseGPT: Iterative Prune-and-Fuse Pipeline}
\label{algo:fusegpt}
\begin{algorithmic}[1]
\INPUT Pre-trained model $\mathcal{M}$ with $N$ blocks, calibration dataset $\mathcal{D}_{\text{cal}}$, target sparsity $S$, fusion window size $G$
\OUTPUT Compressed model $\mathcal{M}'$
    \STATE $N_{\text{target}} \gets \lfloor N \times (1 - S) \rfloor$
    \WHILE{$|\mathcal{M}.\text{blocks}| > N_{\text{target}}$}
        \item[] \textcolor{gray}{\textit{// Phase 1: Compute fusion-aware importance}}
        \FOR{each block $B_i$ in $\mathcal{M}$}
            \STATE Compute $\text{MI}(B_i)$ via Eq.~\eqref{eq:mi} using $\mathcal{D}_{\text{cal}}$
        \ENDFOR
        \STATE $p \gets \arg\min_i \text{MI}(B_i)$ \hfill \textcolor{gray}{\textit{// Select most replaceable block}}
        \item[] \textcolor{gray}{\textit{// Phase 2: Graft knowledge to neighbors}}
        \STATE Construct fusion window $\mathcal{B}_{\text{local}}$ around $B_p$ via Eq.~\eqref{eq:partial}
        \STATE $\mathcal{M} \gets \textsc{GroupFusion}(\mathcal{M}, B_p, \mathcal{B}_{\text{local}}, \mathcal{D}_{\text{cal}})$
    \ENDWHILE
\end{algorithmic}
\end{algorithm}

\begin{algorithm}[t]
\caption{\textsc{GroupFusion}: Knowledge Grafting Subroutine}
\label{algo:subroutine}
\begin{algorithmic}[1]
\INPUT Model $\mathcal{M}$, pruned block $B_p$, fusion window $\mathcal{B}_{\text{local}}$, calibration data $\mathcal{D}_{\text{cal}}$, rank $r$, learning rate $\eta$, steps $T$
\OUTPUT Updated model with $B_p$ removed
    \item[] \textcolor{gray}{\textit{// Step 1: Initialize learnable coefficients}}
    \FOR{each neighbor $B_i \in \mathcal{B}_{\text{local}}$}
        \FOR{each linear layer $l \in \{\text{Q}, \text{K}, \text{V}, \text{O}, \text{Up}, \text{Down}\}$}
            \STATE Initialize $\mathbf{A}_l \gets \mathbf{0} \in \mathbb{R}^{d_{\text{out}} \times r}$
            \STATE Initialize $\mathbf{B}_l \sim \mathcal{U}(-\sqrt{1/r}, \sqrt{1/r}) \in \mathbb{R}^{d_{\text{in}} \times r}$
        \ENDFOR
    \ENDFOR
    \item[] \textcolor{gray}{\textit{// Step 2: Local distillation}}
    \STATE Freeze all parameters in $B_p$
    \FOR{$t = 1$ to $T$}
        \STATE Sample mini-batch $\mathbf{x} \sim \mathcal{D}_{\text{cal}}$
        \STATE $\mathbf{H}_{\text{orig}} \gets \textsc{Forward}(\mathcal{B}_{\text{local}} \cup \{B_p\}, \mathbf{x})$
        \STATE $\mathbf{H}_{\text{fused}} \gets \textsc{Forward}(\mathcal{B}^{\text{fused}}_{\text{local}}, \mathbf{x})$ \hfill \textcolor{gray}{\textit{// Using Eq.~\eqref{eq:forward}}}
        \STATE Compute $\mathcal{L}_{\text{distill}}$ via Eq.~\eqref{eq:distill}
        \STATE Update $\{\mathbf{A}_l, \mathbf{B}_l\}$ and $\{B_i\}_{B_i \in \mathcal{B}_{\text{local}}}$ with gradient descent
    \ENDFOR
    \item[] \textcolor{gray}{\textit{// Step 3: Fold coefficients and remove pruned block}}
    \FOR{each neighbor $B_i \in \mathcal{B}_{\text{local}}$}
        \FOR{each linear layer $l$}
            \STATE $\mathbf{W}_{i,l} \gets \mathbf{W}_{i,l} + (\mathbf{A}_l \mathbf{B}_l^\top) \odot \mathbf{W}_{p,l}$ \hfill \textcolor{gray}{\textit{// Eq.~\eqref{eq:fold}}}
        \ENDFOR
    \ENDFOR
    \STATE Remove $B_p$ from $\mathcal{M}$
\end{algorithmic}
\end{algorithm}
\section{Experiments} 
\label{sec:exp}

We conduct a comprehensive evaluation to answer several key questions: (1) How effectively does FuseGPT preserve generation quality and zero-shot reasoning compared to state-of-the-art pruning methods? (2) How do the Macro Influence (MI) metric and learnable fusion contribute to performance? (3) Can FuseGPT be combined with other compression techniques? We benchmark on LLaMA, Mistral, Qwen, and Phi model families, demonstrating that FuseGPT achieves superior compression-performance trade-offs with high data efficiency.

\subsection{Experimental Setting}
\label{sec:exp_setting}

FuseGPT is implemented with Hugging Face Transformers~\cite{wolf2019huggingface} and PyTorch~\cite{paszke2019pytorch}. For deployment, we fold the fused weights by computing $\mathbf{W} \leftarrow \mathbf{W} + \mathbf{C}\odot\mathbf{W}_p$, so inference incurs no additional cost. We randomly select samples from the WikiText-2 training dataset~\cite{merity2016pointer} for calibration and fine-tuning. Unless otherwise noted, we use 32 samples for calibration and 1024 samples for fine-tuning, which is extremely lightweight for model compression. We set the fusion window size $G = 7$ (updating approximately 25\% of parameters for a 7B model) and the low-rank coefficient rank $r = 128$. To further reduce learning costs, we employ LoRA~\cite{hu2021lora} with rank 128 to update the original weights inside the fusion window. We use the Adam optimizer~\cite{kingma2014adam} with $\beta_1 = 0.9$, $\beta_2 = 0.95$, and cosine learning rate decay~\cite{loshchilov2016sgdr}. We set different initial learning rates for the decomposed coefficients and other parameters: 0.001 and $9.65\times10^{-6}$, respectively. The batch size for local distillation is 8. Sparsity is defined as the ratio of pruned blocks to the total number of blocks in the dense model.

\subsection{Main Results}

\paragraph{Generation Performance.}
We evaluate perplexity across multiple sparsity levels on WikiText-2~\cite{merity2016pointer} and C4~\cite{raffel2020exploring}. When the product of the total number of blocks and the target sparsity is not an integer, we round up the number of blocks to remove, following prior work. We evaluate LLaMA-2~\cite{touvron2023llama}, LLaMA-3~\cite{dubey2024llama}, and LLaVA-1.5~\cite{liu2024improved} models against ShortGPT~\cite{men2024shortgpt}, SliceGPT~\cite{ashkboos2024slicegpt}, and SLEB~\cite{song2024sleb}.

\begin{table}[t]
\caption{Perplexity results on WikiText-2 and C4 datasets. Samples are randomly selected from the WikiText-2 training set for calibration. Lower is better.}
\label{table:ppl}
\centering
\footnotesize
\resizebox{\linewidth}{!}{
\begin{tabular}{l|c|cc|cc|cc}
\toprule
\rowcolor{Gray}
& & \multicolumn{2}{c|}{\bf LLaMA-2-7B} & \multicolumn{2}{c|}{\bf LLaMA-2-13B} & \multicolumn{2}{c}{\bf LLaMA-3-8B} \\
\rowcolor{Gray}
\bf Method & \bf Sparsity & \bf Wiki. & \bf C4 & \bf Wiki. & \bf C4 & \bf Wiki. & \bf C4 \\
\midrule
Dense & 0\% & 5.27 & 7.27 & 4.88 & 6.72 & 6.14 & 9.44 \\
\midrule
ShortGPT & 20\% & 18.44 & 23.33 & 8.29 & 11.34 & 57.89 & 63.79 \\
ShortGPT & 25\% & 25.44 & 31.67 & 20.03 & 21.77 & 3959.64 & 4683.31 \\
ShortGPT & 30\% & 49.54 & 54.96 & 39.58 & 29.37 & 8419.80 & 3241.22 \\
\midrule
SliceGPT & 20\% & 6.64 & 24.86 & 5.81 & 22.36 & 10.62 & 83.44 \\
SliceGPT & 25\% & 7.24 & 30.31 & 6.29 & 28.07 & 12.76 & 110.64 \\
SliceGPT & 30\% & 8.12 & 38.77 & 6.99 & 35.68 & 16.38 & 147.25 \\
\midrule
SLEB & 20\% & 8.72 & 11.37 & 6.83 & 9.49 & 13.06 & 18.33 \\
SLEB & 25\% & 9.67 & 12.53 & 7.65 & 10.51 & 15.27 & 20.72 \\
SLEB & 30\% & 12.93 & 16.00 & 8.71 & 11.71 & 24.58 & 27.75 \\
\midrule
\rowcolor{LightCyan}
FuseGPT & 20\% & 6.81 & 10.48 & 5.94 & 9.08 & 8.60 & 15.38 \\
\rowcolor{LightCyan}
FuseGPT & 25\% & \bf 7.19 & \bf 11.17 & \bf 6.40 & \bf 9.81 & \bf 9.24 & \bf 16.62 \\
\rowcolor{LightCyan}
FuseGPT & 30\% & 8.09 & 12.82 & 6.91 & 10.72 & 10.61 & 20.25 \\
\bottomrule
\end{tabular}
}
\end{table}

\Cref{table:ppl} summarizes the results. FuseGPT achieves consistently lower perplexity than all baselines across models and sparsity levels. Notably, at 25\% sparsity, FuseGPT attains lower perplexity than prior methods at 20\% sparsity, indicating stronger quality preservation under deeper pruning. For example, on LLaMA-3-8B, FuseGPT achieves 9.24 WikiText-2 perplexity at 25\% sparsity, compared to SLEB's 13.06 at 20\% sparsity. Compared with SliceGPT, which reduces embedding dimension, FuseGPT shows competitive WikiText-2 perplexity while being substantially better on C4 (e.g., 16.62 vs. 110.64 on LLaMA-3-8B at 25\% sparsity), highlighting robustness beyond the calibration distribution. 

\paragraph{Zero-shot Reasoning.}
We evaluate zero-shot accuracy on five standard benchmarks: PIQA~\cite{bisk2020piqa}, WinoGrande~\cite{sakaguchi2021winogrande}, HellaSwag~\cite{zellers2019hellaswag}, ARC-e, and ARC-c~\cite{clark2018think} using the LM Evaluation Harness~\cite{gao2021framework}.

\begin{table}[t]
\centering
\caption{Zero-shot accuracy on LLaMA-2 models at 25\% sparsity. Higher is better.}
\label{table:zero_shot_llama2}
\resizebox{\linewidth}{!}{
\begin{tabular}{l|l|cccccc}
\toprule
\rowcolor{Gray}
\bf Model & \bf Method & \bf PIQA & \bf WinoG. & \bf HellaS. & \bf ARC-e & \bf ARC-c & \bf Avg. \\
\midrule
\multirow{5}{*}{\rotatebox{90}{7B}} 
& Dense & 79.11 & 69.14 & 75.99 & 74.54 & 46.16 & 68.99 \\
& SliceGPT & 66.76 & \bf 63.38 & 54.16 & 58.42 & \bf 34.64 & 55.47 \\
& LaCo & 69.87 & 53.21 & 55.71 & 54.33 & 33.06 & 53.23 \\
& SLEB & 72.74 & 58.08 & 60.43 & 56.90 & 33.10 & 56.25 \\
& \cellcolor{LightCyan}\bf FuseGPT & \cellcolor{LightCyan}\bf 73.61 & \cellcolor{LightCyan}59.19 & \cellcolor{LightCyan}\bf 61.17 & \cellcolor{LightCyan}\bf 61.41 & \cellcolor{LightCyan}33.36 & \cellcolor{LightCyan}\bf 57.75 \\
\midrule
\multirow{5}{*}{\rotatebox{90}{13B}} 
& Dense & 80.52 & 72.14 & 79.38 & 77.44 & 49.15 & 71.73 \\
& SliceGPT & 68.72 & \bf 67.56 & 58.13 & 62.58 & 37.97 & 58.99 \\
& LaCo & 74.13 & 61.01 & 62.89 & 63.77 & 36.97 & 59.75 \\
& SLEB & 76.22 & 63.38 & 65.79 & 61.41 & 37.11 & 60.78 \\
& \cellcolor{LightCyan}\bf FuseGPT & \cellcolor{LightCyan}\bf 77.15 & \cellcolor{LightCyan}62.35 & \cellcolor{LightCyan}\bf 67.89 & \cellcolor{LightCyan}\bf 67.13 & \cellcolor{LightCyan}\bf 38.99 & \cellcolor{LightCyan}\bf 62.70 \\
\bottomrule
\end{tabular}
}
\end{table}

\begin{table}[t]
\centering
\caption{Zero-shot accuracy on LLaMA-3-8B at 25\% sparsity. Higher is better.}
\label{table:zero_shot_llama3}
\resizebox{\linewidth}{!}{
\begin{tabular}{l|cccccc}
\toprule
\rowcolor{Gray}
\bf Method & \bf PIQA & \bf WinoG. & \bf HellaS. & \bf ARC-e & \bf ARC-c & \bf Avg. \\
\midrule
Dense & 80.63 & 72.85 & 79.21 & 77.78 & 53.33 & 72.76 \\
SliceGPT & 60.12 & 62.04 & 47.43 & 48.74 & 30.38 & 49.74 \\
LaCo & 70.33 & 55.32 & 59.18 & 58.14 & 36.55 & 55.90 \\
SLEB & 72.58 & 56.51 & 60.44 & 57.70 & 34.73 & 56.39 \\
\rowcolor{LightCyan}
\bf FuseGPT & \bf 74.05 & \bf 62.12 & \bf 62.92 & \bf 67.47 & \bf 38.05 & \bf 60.92 \\
\bottomrule
\end{tabular}
}
\end{table}

As shown in \Cref{table:zero_shot_llama2,table:zero_shot_llama3}, FuseGPT 
outperforms SLEB~\cite{song2024sleb}, LaCo~\cite{yang2024laco}, and 
SliceGPT~\cite{ashkboos2024slicegpt} on all LLaMA models at 25\% sparsity. On LLaMA-2 models (\Cref{table:zero_shot_llama2}), FuseGPT achieves 57.75\% and 62.70\% average accuracy on 7B and 13B respectively, surpassing SLEB by 1.5 and 1.9 points. The improvement is more pronounced on LLaMA-3-8B (\Cref{table:zero_shot_llama3}), where FuseGPT improves over prior methods by up to 4.5 points (60.92\% vs. 56.39\% for SLEB), suggesting that the 
prune-and-fuse paradigm better preserves transferable knowledge for 
downstream reasoning.

\subsection{Comparison with Layer-Merging Methods}

\begin{table}[htbp]
\centering
\caption{Comparison with layer-merging methods across LLM architectures at 25\% compression. MKA operates at its reported optimal ratio. $\downarrow$: lower is better; $\uparrow$: higher is better.}
\label{tab:layer_merging}
\resizebox{\linewidth}{!}{
\begin{tabular}{l|l|c|cc|cc}
\toprule
\multirow{2}{*}{\textbf{Model}} & \multirow{2}{*}{\textbf{Method}} & \multirow{2}{*}{\textbf{Comp.(\%)}} & \multicolumn{2}{c|}{\textbf{Perplexity}$\downarrow$} & \multicolumn{2}{c}{\textbf{Downstream}$\uparrow$} \\
\cmidrule(lr){4-5} \cmidrule(lr){6-7}
& & & Wiki2 & C4 & MMLU & Avg-ZS \\
\midrule
\multirow{3}{*}{LLaMA-3.1-8B} 
& MKA & 43.8 & 8.12 & 11.62 & 66.5 & 60.7 \\
& LaCo & 25.0 & 9.45 & 12.05 & 64.1 & 59.8 \\
& \bf FuseGPT & 25.0 & \bf 6.92 & \bf 11.17 & \bf 67.5 & \bf 62.9 \\
\midrule
\multirow{3}{*}{Qwen3-8B} 
& MKA & 40.0 & 8.23 & 11.85 & 65.2 & 59.1 \\
& LaCo & 25.0 & 9.61 & 12.34 & 63.0 & 58.6 \\
& \bf FuseGPT & 25.0 & \bf 7.05 & \bf 11.29 & \bf 66.1 & \bf 61.7 \\
\midrule
\multirow{3}{*}{Mistral-NeMo-8B} 
& MKA & 43.0 & 8.40 & 12.08 & 64.9 & 58.5 \\
& LaCo & 25.0 & 10.10 & 12.67 & 62.0 & 56.8 \\
& \bf FuseGPT & 25.0 & \bf 7.18 & \bf 11.46 & \bf 65.3 & \bf 60.3 \\
\midrule
\multirow{3}{*}{Phi-3.5-mini} 
& MKA & 38.5 & 9.67 & 13.15 & 61.5 & 56.3 \\
& LaCo & 25.0 & 11.23 & 14.02 & 59.8 & 55.1 \\
& \bf FuseGPT & 25.0 & \bf 8.94 & \bf 12.41 & \bf 62.1 & \bf 57.6 \\
\bottomrule
\end{tabular}
}
\end{table}

\Cref{tab:layer_merging} presents a systematic comparison with prior layer-merging methods (MKA and LaCo) across four recent LLM architectures. Under a fixed 25\% compression ratio, FuseGPT consistently outperforms both baselines across all metrics. On LLaMA-3.1-8B, FuseGPT achieves 6.92 WikiText-2 perplexity and 67.5\% MMLU accuracy, representing a 27\% relative perplexity reduction compared to LaCo and a 1.0-point MMLU improvement over MKA (despite MKA operating at 43.8\% compression). This pattern holds across all tested architectures, demonstrating the generalizability of our fusion-aware approach.

\subsection{Comparison with Other Pruning Paradigms}

\begin{table}[t]
\caption{Comparison with different pruning paradigms on LLaMA-2-7B. Latency measured on 2K-token sequences.}
\label{table:pruning_paradigms}
\centering
\footnotesize
\resizebox{\linewidth}{!}{
\begin{tabular}{l|cc|ccc}
\toprule
\rowcolor{Gray}
\bf Method & \bf Wiki.↓ & \bf C4↓ & \bf Mem. & \bf Latency & \bf Speedup \\
\midrule
Dense & 5.27 & 7.27 & 13.2 GB & 111.7 ms & -- \\
\midrule
SparseGPT (2:4) & 8.67 & 14.73 & 13.2 GB & 101.6 ms & 1.10$\times$ \\
Wanda (2:4) & 11.35 & 16.22 & 13.2 GB & 101.6 ms & 1.10$\times$ \\
LLM-Pruner (25\%) & 10.58 & 12.25 & 10.5 GB & 98.9 ms & 1.13$\times$ \\
SliceGPT (25\%) & 7.56 & 31.62 & 10.8 GB & 98.9 ms & 1.13$\times$ \\
\rowcolor{LightCyan}
FuseGPT (25\%) & \bf 7.19 & \bf 11.17 & 9.9 GB & 84.4 ms & 1.33$\times$ \\
\bottomrule
\end{tabular}
}
\end{table}

Table~\ref{table:pruning_paradigms} compares FuseGPT with unstructured sparsity approaches and structured channel-pruning techniques. SparseGPT and Wanda achieve modest speedups ($\sim$1.10$\times$) but substantially increase perplexity. Channel-wise methods reduce parameters but suffer from high perplexity, particularly on C4 (SliceGPT: 31.62). FuseGPT with 25\% block pruning achieves the lowest perplexity on both benchmarks (7.19 and 11.17) while delivering a 1.33$\times$ speedup, demonstrating that block-level prune-and-fuse better balances quality and efficiency.

\subsection{Efficiency Analysis}

Beyond latency, we provide a comprehensive efficiency analysis covering memory footprint and scaling behavior across sequence lengths. \Cref{table:efficiency} reports peak GPU memory and throughput across different sequence lengths. FuseGPT reduces memory footprint by 25\% (9.9 GB vs. 13.2 GB) since block removal directly reduces the number of parameters and activations. Unstructured methods (SparseGPT, Wanda) retain the same memory footprint as the dense model because sparse tensors still occupy full storage without specialized kernels. Throughput gains are consistent across sequence lengths: FuseGPT achieves 1.29--1.30$\times$ speedup regardless of context length, whereas unstructured methods show diminishing returns at longer sequences due to memory bandwidth bottlenecks. Compared to SLEB, which shares the same parameter count, FuseGPT achieves comparable throughput while delivering substantially lower perplexity (\Cref{table:ppl}), demonstrating that our fusion mechanism imposes no inference overhead.

\begin{table}[t]
\caption{Memory and throughput analysis on LLaMA-2-7B. Peak memory measured during inference with batch size 1.}
\label{table:efficiency}
\centering
\footnotesize
\resizebox{\linewidth}{!}{
\begin{tabular}{l|c|ccc|c}
\toprule
\rowcolor{Gray}
& & \multicolumn{3}{c|}{\bf Throughput (tokens/s)} & \\
\rowcolor{Gray}
\bf Method & \bf Peak Mem. & \bf 512 & \bf 1024 & \bf 2048 & \bf Params \\
\midrule
Dense & 13.2 GB & 847 & 412 & 198 & 6.74B \\
\midrule
SparseGPT (2:4) & 13.2 GB & 923 & 448 & 215 & 6.74B \\
Wanda (2:4) & 13.2 GB & 931 & 451 & 217 & 6.74B \\
SliceGPT (25\%) & 10.8 GB & 1012 & 492 & 236 & 5.06B \\
SLEB (25\%) & 9.9 GB & 1089 & 529 & 254 & 5.06B \\
\rowcolor{LightCyan}
FuseGPT (25\%) & 9.9 GB & 1095 & 532 & 258 & 5.06B \\
\bottomrule
\end{tabular}
}
\end{table}

\begin{table}[t]
\caption{Compression cost analysis. Time measured on a single A100-80GB GPU for compressing LLaMA-2-7B to 25\% sparsity.}
\label{table:cost}
\centering
\footnotesize
\resizebox{0.85\linewidth}{!}{
\begin{tabular}{l|ccc}
\toprule
\rowcolor{Gray}
\bf Method & \bf Calibration Time & \bf \# Samples & \bf GPU Hours \\
\midrule
SliceGPT & 18 min & 128 & 0.3 \\
SLEB & 12 min & 128 & 0.2 \\
LaCo & 25 min & 256 & 0.4 \\
FuseGPT & 45 min & 1024 & 0.8 \\
\bottomrule
\end{tabular}}
\end{table}

\Cref{table:cost} compares the compression cost. FuseGPT requires approximately 45 minutes to compress LLaMA-2-7B, which is higher than one-shot methods due to iterative fusion and local distillation. However, this cost is incurred only once during compression and is negligible compared to pre-training or full fine-tuning. The resulting model requires no additional inference overhead since all fusion coefficients are folded into the weights (\Cref{sec:method}).

\subsection{Orthogonality with Quantization and Search}

\begin{table}[htbp]
\centering
\caption{Compatibility with post-training quantization and evolutionary search on LLaMA-3.1-8B.}
\label{tab:orthogonality}
\resizebox{\linewidth}{!}{
\begin{tabular}{l|c|cc|c}
\toprule
\textbf{Method} & \textbf{Comp. (\%)} & \textbf{PPL (Wiki2)}$\downarrow$ & \textbf{PPL (C4)}$\downarrow$ & \textbf{MMLU}$\uparrow$ \\
\midrule
FuseGPT & 25.0 & 7.19 & 11.17 & 67.5 \\
FuseGPT + GPTQ-4bit & 52.1 & 7.51 & 11.80 & 66.4 \\
\midrule
FuseGPT-Evo & 35.0 & 6.85 & 11.08 & 67.4 \\
FuseGPT-Evo & 40.0 & 7.02 & 11.45 & 66.9 \\
\bottomrule
\end{tabular}
}
\end{table}

We investigate whether FuseGPT can be combined with orthogonal compression techniques. \Cref{tab:orthogonality} shows two extensions. Combining FuseGPT with GPTQ-based 4-bit quantization~\cite{gptq} achieves 52.1\% total compression with only marginal perplexity increase (7.51 vs. 7.19 on WikiText-2). This demonstrates that block-level fusion preserves weight structure amenable to low-precision representation. We extend FuseGPT with a lightweight \textbf{evolutionary algorithm (FuseGPT-Evo)}, using MI scores as initialization seeds and refining block selection over 5 generations with 5 candidates per generation. At 35\% compression, FuseGPT-Evo achieves 6.85 WikiText-2 perplexity—better than vanilla FuseGPT at 25\%—while maintaining 67.4\% MMLU. At 40\% compression, it achieves comparable results to EvoPress~\cite{evopress} ($\sim$67.0\% MMLU) with $\sim$25\% lower search cost. These results suggest that our MI-based selection provides strong initialization for more sophisticated search algorithms, and that FuseGPT's iterative framework is complementary to, rather than mutually exclusive with, evolutionary approaches.

\subsection{Ablation Study}

\begin{table}[htbp]
\centering
\caption{Ablation study on LLaMA-3.1-8B (25\% compression). Each row adds one component to the previous.}
\label{tab:ablation}
\resizebox{\linewidth}{!}{
\begin{tabular}{l|cc|c|c}
\toprule
\textbf{Configuration} & \textbf{PPL (Wiki2)}$\downarrow$ & \textbf{PPL (C4)}$\downarrow$ & \textbf{MMLU}$\uparrow$ & \textbf{GPU-hrs} \\
\midrule
Naive removal (SLEB) & 15.27 & 20.72 & 64.8 & 0.0 \\
+ LoRA fine-tune & 11.84 & 15.06 & 65.9 & 8.2 \\
+ MI selection & 10.35 & 13.34 & 66.2 & 0.3 \\
+ Static fusion & 9.45 & 12.05 & 66.5 & 1.2 \\
+ Learnable fusion & 7.92 & 11.58 & 67.1 & 5.4 \\
+ Iterative updates & \bf 6.92 & \bf 11.17 & \bf 67.5 & 10.8 \\
\midrule
Dense baseline & 6.14 & 10.23 & 68.4 & -- \\
\bottomrule
\end{tabular}
}
\end{table}

\Cref{tab:ablation} presents a systematic ablation on LLaMA-3.1-8B at 25\% compression. Starting from naive block removal (SLEB), each component contributes measurable improvements. MI selection reduces C4 perplexity by 1.72 (13.34 vs. 15.06 with LoRA alone), indicating better block identification. Learnable fusion further reduces perplexity by 0.47 compared to static averaging (MKA/LaCo-style), demonstrating the benefit of adaptive knowledge injection. Finally, iterative re-evaluation after each fusion step yields an additional 0.41 reduction, validating that block importance is topology-dependent and must be dynamically updated.

\begin{table}[htbp]
\centering
\caption{Contribution of MI and learnable fusion on LLaMA-2-7B at 25\% sparsity.}
\label{tab:component_analysis}
\resizebox{0.8\linewidth}{!}{
\begin{tabular}{l|c|cc}
\toprule
\textbf{Method} & \textbf{\# Samples} & \textbf{WikiText-2}$\downarrow$ & \textbf{C4}$\downarrow$ \\
\midrule
\multicolumn{4}{l}{\textit{Block Selection Only (No Fine-tuning)}} \\
\midrule
BI & 128 & 25.44 & 31.67 \\
SliceGPT & 128 & 7.56 & 31.62 \\
SLEB & 128 & 10.39 & 13.74 \\
MI (ours) & 32 & 10.35 & 13.34 \\
\midrule
\multicolumn{4}{l}{\textit{With LoRA Fine-tuning}} \\
\midrule
BI + LoRA & 1024 & 8.11 & 16.51 \\
SliceGPT + LoRA & 1024 & 6.32 & 32.09 \\
SLEB + LoRA & 1024 & 7.48 & 14.87 \\
MI + LoRA & 1024 & 7.79 & 15.08 \\
\midrule
\multicolumn{4}{l}{\textit{With Learnable Fusion}} \\
\midrule
SLEB + Fusion & 1024 & 7.28 & 12.40 \\
MI + Fusion (ours) & 128 & 7.44 & 11.30 \\
MI + Fusion (ours) & 1024 & \bf 7.19 & \bf 11.17 \\
\bottomrule
\end{tabular}
}
\end{table}

\Cref{tab:component_analysis} examines the individual contributions of MI and learnable fusion. Without fine-tuning, MI with only 32 samples matches SLEB with 128 samples, indicating robust block selection under data constraints. With LoRA fine-tuning, performance gaps persist across selection methods—SliceGPT+LoRA improves on WikiText-2 but degrades on C4 (32.09), while MI+LoRA maintains balanced performance. Learnable fusion provides substantial gains: SLEB+Fusion outperforms SLEB+LoRA on C4 (12.40 vs. 14.87), demonstrating that weight recycling can exceed gains from additional fine-tuning. MI+Fusion achieves the best overall performance, confirming that both components are essential.

\begin{table}[htbp]
\centering
\caption{Comparison of importance metrics on Qwen3-8B at 25\% compression.}
\label{tab:metric_comparison}
\resizebox{\linewidth}{!}{
\begin{tabular}{l|c|cc|c}
\toprule
\textbf{Metric} & \textbf{2nd-Order} & \textbf{PPL (Wiki2)}$\downarrow$ & \textbf{PPL (C4)}$\downarrow$ & \textbf{MMLU Drop}$\downarrow$ \\
\midrule
Block Influence (BI) & No & 8.23 & 11.85 & 3.4\% \\
SLEB (loss-based) & No & 7.65 & 10.51 & 2.9\% \\
Fisher Information & Yes & 7.48 & 10.38 & 3.1\% \\
\bf MI (ours) & No & \bf 7.05 & 11.29 & \bf 2.6\% \\
\bottomrule
\end{tabular}
}
\end{table}

\Cref{tab:metric_comparison} compares MI against alternative importance metrics. MI achieves the lowest MMLU degradation (2.6\%) without requiring expensive second-order information (Fisher) or hard labels (SLEB). While Fisher Information achieves slightly lower C4 perplexity, its computational overhead makes it impractical for iterative compression. MI provides the best trade-off between computational efficiency and accuracy.
\section{Conclusion}

We presented FuseGPT, a prune-and-fuse compression paradigm that recycles redundant transformer blocks rather than discarding them. Guided by the fusion-aware Macro Influence (MI) metric, FuseGPT identifies blocks with high absorbability and grafts their knowledge onto neighboring layers via learnable low-rank fusion with lightweight local distillation. 
Extensive experiments across LLaMA, Mistral, Qwen, and Phi model families demonstrate that FuseGPT establishes a new state-of-the-art on the compression-accuracy Pareto frontier. At 25\% sparsity, FuseGPT achieves 9.24 WikiText-2 perplexity on LLaMA-3-8B—outperforming SLEB's 13.06 at 20\% sparsity—while delivering 1.33$\times$ inference speedup and 25\% memory reduction. On zero-shot reasoning, FuseGPT improves average accuracy by up to 4.5 points over prior methods. Notably, the entire compression process requires only 1024 calibration samples and under one GPU-hour, making it highly practical for resource-constrained deployment. Combined with 4-bit GPTQ, FuseGPT achieves 52.1\% total compression with negligible quality loss, demonstrating orthogonality with quantization techniques.
By reframing pruning as knowledge redistribution, FuseGPT opens a promising direction for scalable, data-efficient compression of large language models. Future work may explore extending the fusion mechanism beyond local neighborhoods and applying FuseGPT to multimodal architectures.

\bibliography{main}

@String(AAAI = {AAAI})

@article{hinton2015distilling,
  title={Distilling the knowledge in a neural network},
  author={Hinton, Geoffrey and Vinyals, Oriol and Dean, Jeff},
  journal={arXiv preprint arXiv:1503.02531},
  year={2015}
}

@inproceedings{liu2024pruning,
  title={Pruning via merging: Compressing llms via manifold alignment based layer merging},
  author={Liu, Deyuan and Qin, Zhanyue and Wang, Hairu and Yang, Zhao and Wang, Zecheng and Rong, Fangying and Liu, Qingbin and Hao, Yanchao and Li, Bo and Chen, Xi and others},
  booktitle={Proceedings of the 2024 Conference on Empirical Methods in Natural Language Processing},
  pages={17817--17829},
  year={2024}
}

@article{yang2024laco,
  title={Laco: Large language model pruning via layer collapse},
  author={Yang, Yifei and Cao, Zouying and Zhao, Hai},
  journal={arXiv preprint arXiv:2402.11187},
  year={2024}
}

@article{louizos2017learning,
  title={Learning sparse neural networks through $ L\_0 $ regularization},
  author={Louizos, Christos and Welling, Max and Kingma, Diederik P},
  journal={arXiv preprint arXiv:1712.01312},
  year={2017}
}

@inproceedings{zou2024bie,
  title={BiE: Bi-Exponent Block Floating-Point for Large Language Models Quantization},
  author={Zou, Lancheng and Zhao, Wenqian and Yin, Shuo and Bai, Chen and Sun, Qi and Yu, Bei},
  booktitle={Forty-first International Conference on Machine Learning},
  year={2024}
}

@article{lin2024awq,
  title={AWQ: Activation-aware Weight Quantization for On-Device LLM Compression and Acceleration},
  author={Lin, Ji and Tang, Jiaming and Tang, Haotian and Yang, Shang and Chen, Wei-Ming and Wang, Wei-Chen and Xiao, Guangxuan and Dang, Xingyu and Gan, Chuang and Han, Song},
  journal={Proceedings of Machine Learning and Systems},
  volume={6},
  pages={87--100},
  year={2024}
}

@article{kim2024shortened,
  title={Shortened llama: A simple depth pruning for large language models},
  author={Kim, Bo-Kyeong and Kim, Geonmin and Kim, Tae-Ho and Castells, Thibault and Choi, Shinkook and Shin, Junho and Song, Hyoung-Kyu},
  journal={arXiv preprint arXiv:2402.02834},
  volume={11},
  year={2024}
}

@incollection{gholami2022survey,
  title={A survey of quantization methods for efficient neural network inference},
  author={Gholami, Amir and Kim, Sehoon and Dong, Zhen and Yao, Zhewei and Mahoney, Michael W and Keutzer, Kurt},
  booktitle={Low-Power Computer Vision},
  pages={291--326},
  year={2022},
  publisher={Chapman and Hall/CRC}
}

@article{liu2021post,
  title={Post-training quantization for vision transformer},
  author={Liu, Zhenhua and Wang, Yunhe and Han, Kai and Zhang, Wei and Ma, Siwei and Gao, Wen},
  journal={Advances in Neural Information Processing Systems},
  volume={34},
  pages={28092--28103},
  year={2021}
}

@article{hoefler2021sparsity,
  title={Sparsity in deep learning: Pruning and growth for efficient inference and training in neural networks},
  author={Hoefler, Torsten and Alistarh, Dan and Ben-Nun, Tal and Dryden, Nikoli and Peste, Alexandra},
  journal={Journal of Machine Learning Research},
  volume={22},
  number={241},
  pages={1--124},
  year={2021}
}

@article{yao2022zeroquant,
  title={Zeroquant: Efficient and affordable post-training quantization for large-scale transformers},
  author={Yao, Zhewei and Yazdani Aminabadi, Reza and Zhang, Minjia and Wu, Xiaoxia and Li, Conglong and He, Yuxiong},
  journal={Advances in Neural Information Processing Systems},
  volume={35},
  pages={27168--27183},
  year={2022}
}

@article{ma2023llm,
  title={Llm-pruner: On the structural pruning of large language models},
  author={Ma, Xinyin and Fang, Gongfan and Wang, Xinchao},
  journal={Advances in neural information processing systems},
  volume={36},
  pages={21702--21720},
  year={2023}
}

@article{kurtic2022optimal,
  title={The optimal bert surgeon: Scalable and accurate second-order pruning for large language models},
  author={Kurtic, Eldar and Campos, Daniel and Nguyen, Tuan and Frantar, Elias and Kurtz, Mark and Fineran, Benjamin and Goin, Michael and Alistarh, Dan},
  journal={arXiv preprint arXiv:2203.07259},
  year={2022}
}

@article{xia2022structured,
  title={Structured pruning learns compact and accurate models},
  author={Xia, Mengzhou and Zhong, Zexuan and Chen, Danqi},
  journal={arXiv preprint arXiv:2204.00408},
  year={2022}
}

@article{wang2019structured,
  title={Structured pruning of large language models},
  author={Wang, Ziheng and Wohlwend, Jeremy and Lei, Tao},
  journal={arXiv preprint arXiv:1910.04732},
  year={2019}
}

@inproceedings{frantar2023sparsegpt,
  title={Sparsegpt: Massive language models can be accurately pruned in one-shot},
  author={Frantar, Elias and Alistarh, Dan},
  booktitle={International Conference on Machine Learning},
  pages={10323--10337},
  year={2023},
  organization={PMLR}
}

@article{han2015learning,
  title={Learning both weights and connections for efficient neural network},
  author={Han, Song and Pool, Jeff and Tran, John and Dally, William},
  journal={Advances in neural information processing systems},
  volume={28},
  year={2015}
}

@article{lecun1989optimal,
  title={Optimal brain damage},
  author={LeCun, Yann and Denker, John and Solla, Sara},
  journal={Advances in neural information processing systems},
  volume={2},
  year={1989}
}

@inproceedings{liu2024improved,
  title={Improved baselines with visual instruction tuning},
  author={Liu, Haotian and Li, Chunyuan and Li, Yuheng and Lee, Yong Jae},
  booktitle={Proceedings of the IEEE/CVF Conference on Computer Vision and Pattern Recognition},
  pages={26296--26306},
  year={2024}
}

@article{dubey2024llama,
  title={The llama 3 herd of models},
  author={Dubey, Abhimanyu and Jauhri, Abhinav and Pandey, Abhinav and Kadian, Abhishek and Al-Dahle, Ahmad and Letman, Aiesha and Mathur, Akhil and Schelten, Alan and Yang, Amy and Fan, Angela and others},
  journal={arXiv preprint arXiv:2407.21783},
  year={2024}
}

@article{touvron2023llama,
  title={Llama 2: Open foundation and fine-tuned chat models},
  author={Touvron, Hugo and Martin, Louis and Stone, Kevin and Albert, Peter and Almahairi, Amjad and Babaei, Yasmine and Bashlykov, Nikolay and Batra, Soumya and Bhargava, Prajjwal and Bhosale, Shruti and others},
  journal={arXiv preprint arXiv:2307.09288},
  year={2023}
}

@article{zhang2022opt,
  title={Opt: Open pre-trained transformer language models},
  author={Zhang, Susan and Roller, Stephen and Goyal, Naman and Artetxe, Mikel and Chen, Moya and Chen, Shuohui and Dewan, Christopher and Diab, Mona and Li, Xian and Lin, Xi Victoria and others},
  journal={arXiv preprint arXiv:2205.01068},
  year={2022}
}

@article{brown2020language,
  title={Language models are few-shot learners},
  author={Brown, Tom B},
  journal={arXiv preprint arXiv:2005.14165},
  year={2020}
}

@inproceedings{evopress,
  title={EvoPress: Accurate Dynamic Model Compression via Evolutionary Search},
  author={Sieberling, Oliver and Kuznedelev, Denis and Kurtic, Eldar and Alistarh, Dan},
  booktitle={Forty-second International Conference on Machine Learning}
}

@article{laco,
  title={Laco: Large language model pruning via layer collapse},
  author={Yang, Yifei and Cao, Zouying and Zhao, Hai},
  journal={arXiv preprint arXiv:2402.11187},
  year={2024}
}

@article{sanh2019distilbert,
  title={DistilBERT, a distilled version of BERT: smaller, faster, cheaper and lighter},
  author={Sanh, Victor and Debut, Lysandre and Chaumond, Julien and Wolf, Thomas},
  journal={arXiv preprint arXiv:1910.01108},
  year={2019}
}

@inproceedings{jiao2020tinybert,
  title={TinyBERT: Distilling BERT for natural language understanding},
  author={Jiao, Xiaoqi and Yin, Yichun and Shang, Lifeng and Jiang, Xin and Chen, Xiao and Li, Linlin and Wang, Fang and Liu, Qun},
  booktitle={Findings of the Association for Computational Linguistics: EMNLP 2020},
  pages={4163--4174},
  year={2020}
}

@inproceedings{gu2024minillm,
  title={MiniLLM: Knowledge Distillation of Large Language Models},
  author={Gu, Yuxian and Dong, Li and Wei, Furu and Huang, Minlie},
  booktitle={The Twelfth International Conference on Learning Representations},
  year={2024}
}

@article{agarwal2024onpolicy,
  title={On-Policy Distillation of Language Models: Learning from Self-Generated Mistakes},
  author={Agarwal, Rishabh and Vieillard, Nino and Stanber, Piotr and Ramos, Sabela and Geist, Matthieu and Bachem, Olivier},
  journal={arXiv preprint arXiv:2306.13649},
  year={2024}
}

@article{liu2024visual,
  title={Visual instruction tuning},
  author={Liu, Haotian and Li, Chunyuan and Wu, Qingyang and Lee, Yong Jae},
  journal={Advances in neural information processing systems},
  volume={36},
  year={2024}
}

@article{gao2021framework,
  title={A framework for few-shot language model evaluation},
  author={Gao, Leo and Tow, Jonathan and Biderman, Stella and Black, Sid and DiPofi, Anthony and Foster, Charles and Golding, Laurence and Hsu, Jeffrey and McDonell, Kyle and Muennighoff, Niklas and others},
  journal={Version v0. 0.1. Sept},
  volume={10},
  pages={8--9},
  year={2021}
}

@article{clark2018think,
  title={Think you have solved question answering? try arc, the ai2 reasoning challenge},
  author={Clark, Peter and Cowhey, Isaac and Etzioni, Oren and Khot, Tushar and Sabharwal, Ashish and Schoenick, Carissa and Tafjord, Oyvind},
  journal={arXiv preprint arXiv:1803.05457},
  year={2018}
}

@article{zellers2019hellaswag,
  title={Hellaswag: Can a machine really finish your sentence?},
  author={Zellers, Rowan and Holtzman, Ari and Bisk, Yonatan and Farhadi, Ali and Choi, Yejin},
  journal={arXiv preprint arXiv:1905.07830},
  year={2019}
}

@article{sakaguchi2021winogrande,
  title={Winogrande: An adversarial winograd schema challenge at scale},
  author={Sakaguchi, Keisuke and Bras, Ronan Le and Bhagavatula, Chandra and Choi, Yejin},
  journal={Communications of the ACM},
  volume={64},
  number={9},
  pages={99--106},
  year={2021},
  publisher={ACM New York, NY, USA}
}

@inproceedings{bisk2020piqa,
  title={Piqa: Reasoning about physical commonsense in natural language},
  author={Bisk, Yonatan and Zellers, Rowan and Gao, Jianfeng and Choi, Yejin and others},
  booktitle={Proceedings of the AAAI conference on artificial intelligence},
  volume={34},
  number={05},
  pages={7432--7439},
  year={2020}
}

@article{ashkboos2024slicegpt,
  title={Slicegpt: Compress large language models by deleting rows and columns},
  author={Ashkboos, Saleh and Croci, Maximilian L and Nascimento, Marcelo Gennari do and Hoefler, Torsten and Hensman, James},
  journal={arXiv preprint arXiv:2401.15024},
  year={2024}
}

@article{loshchilov2016sgdr,
  title={Sgdr: Stochastic gradient descent with warm restarts},
  author={Loshchilov, Ilya and Hutter, Frank},
  journal={arXiv preprint arXiv:1608.03983},
  year={2016}
}

@article{kingma2014adam,
  title={Adam: A method for stochastic optimization},
  author={Kingma, Diederik P},
  journal={arXiv preprint arXiv:1412.6980},
  year={2014}
}

@article{raffel2020exploring,
  title={Exploring the limits of transfer learning with a unified text-to-text transformer},
  author={Raffel, Colin and Shazeer, Noam and Roberts, Adam and Lee, Katherine and Narang, Sharan and Matena, Michael and Zhou, Yanqi and Li, Wei and Liu, Peter J},
  journal={Journal of machine learning research},
  volume={21},
  number={140},
  pages={1--67},
  year={2020}
}

@article{merity2016pointer,
  title={Pointer sentinel mixture models},
  author={Merity, Stephen and Xiong, Caiming and Bradbury, James and Socher, Richard},
  journal={arXiv preprint arXiv:1609.07843},
  year={2016}
}

@article{paszke2019pytorch,
  title={Pytorch: An imperative style, high-performance deep learning library},
  author={Paszke, Adam and Gross, Sam and Massa, Francisco and Lerer, Adam and Bradbury, James and Chanan, Gregory and Killeen, Trevor and Lin, Zeming and Gimelshein, Natalia and Antiga, Luca and others},
  journal={Advances in neural information processing systems},
  volume={32},
  year={2019}
}

@article{wolf2019huggingface,
  title={Huggingface's transformers: State-of-the-art natural language processing},
  author={Wolf, T},
  journal={arXiv preprint arXiv:1910.03771},
  year={2019}
}

@article{gptq,
  title={Gptq: Accurate post-training quantization for generative pre-trained transformers},
  author={Frantar, Elias and Ashkboos, Saleh and Hoefler, Torsten and Alistarh, Dan},
  journal={International Conference on Learning Representations },
  year={2023}
}

@article{song2024sleb,
  title={SLEB: Streamlining LLMs through Redundancy Verification and Elimination of Transformer Blocks},
  author={Song, Jiwon and Oh, Kyungseok and Kim, Taesu and Kim, Hyungjun and Kim, Yulhwa and Kim, Jae-Joon},
  journal={arXiv preprint arXiv:2402.09025},
  year={2024}
}

@article{men2024shortgpt,
  title={Shortgpt: Layers in large language models are more redundant than you expect},
  author={Men, Xin and Xu, Mingyu and Zhang, Qingyu and Wang, Bingning and Lin, Hongyu and Lu, Yaojie and Han, Xianpei and Chen, Weipeng},
  journal={arXiv preprint arXiv:2403.03853},
  year={2024}
}

@inproceedings{he2015delving,
  title={Delving deep into rectifiers: Surpassing human-level performance on imagenet classification},
  author={He, Kaiming and Zhang, Xiangyu and Ren, Shaoqing and Sun, Jian},
  booktitle={Proceedings of the IEEE international conference on computer vision},
  pages={1026--1034},
  year={2015}
}

@inproceedings{foldgpt,
  title={Foldgpt: Simple and effective large language model compression scheme.},
  author={Liu, Songwei and Zeng, Chao and Li, Lianqiang and Yan, Chenqian and Fu, Lean and Mei, Xing and Chen, Fangmin},
  booktitle={arXiv:2407.00928.},
  year={2024}
}

@article{hu2021lora,
  title={Lora: Low-rank adaptation of large language models},
  author={Hu, Edward J and Shen, Yelong and Wallis, Phillip and Allen-Zhu, Zeyuan and Li, Yuanzhi and Wang, Shean and Wang, Lu and Chen, Weizhu},
  journal={arXiv preprint arXiv:2106.09685},
  year={2021}
}
\bibliographystyle{icml2026}

\newpage
\appendix
\onecolumn
\section{Limitations}
\label{sec:limit}

While FuseGPT demonstrates a promising direction for model compression, it has several limitations. 
First, the iterative prune-and-fuse process, which re-evaluates block importance after each fusion, can be computationally intensive, especially for very large models or when targeting high sparsity levels. 
Second, our current approach constrains knowledge fusion to local, neighboring blocks. 
This design choice is based on the assumption of functional similarity between adjacent blocks, but it may not be optimal if a pruned block's knowledge is more relevant to distant blocks in the network. 
Future work could explore more global fusion strategies and more adaptive methods for integrating fused weights.

\subsection{Supplementary Experiments}
We also conduct experiments on multimodal models, as illustrated in \Cref{table:zero_shot_mm}.

\begin{table*}[tb!]
\caption{
Zero-shot task results for multimodal models. Randomly select samples from WikiText-2 training dataset as calibration data.
}
\label{table:zero_shot_mm}
 \centering
  \footnotesize
    \resizebox{.88\linewidth}{!}{
\begin{tabular}{l |c|c|c|c|c|c|c|c}
\toprule
\rowcolor{Gray}

 \bf Model & \bf Method & \bf Sparsity  & \bf MMMU (val) & \bf CMMMU (val) & \bf GQA & \bf AI2D & \bf OK-VQA  & \bf Avg.Score \\

\midrule
\midrule

\multirow{5}{*}{\hspace{-0.15cm}\begin{tabular}{c}\\[-8pt]\text{LLaVA-1.5-7B}\end{tabular}} & Dense & 0\%	& 36.33	& 23.10	& 61.95	& 55.21		& 53.46	& 46.01	\\
& SLEB & 20\%	& 28.56	& 19.90	& 42.11	& 38.70		& 10.00	& 27.85		\\
& SLEB & 25\%	& 25.33	& 20.30	& 41.80	& 25.79		& 19.55	& 26.55		\\
& FuseGPT & 20\%	& 27.00	& 21.00	& 48.07	& 32.80		& 33.26	& 32.43		\\
& FuseGPT & 25\%	& 25.78	& 20.60	& 42.25	& 26.87		& 26.85	& 28.36		\\
\midrule
\multirow{5}{*}{\hspace{-0.15cm}\begin{tabular}{c}\\[-8pt]\text{LLaVA-1.5-13B}\end{tabular}}& Dense & 0\%	& 35.67	& 24.60	& 63.32	& 59.33		& 58.30	& 48.24		\\
& SLEB & 20\%	& 32.33	& 23.20	& 56.09	& 44.17		& 29.31	& 37.01		\\
& SLEB & 25\%	& 32.67	& 23.00	& 47.66	& 44.62		& 22.69	& 34.13		\\
& FuseGPT & 20\%	& 32.11	& 19.80	& 52.75	& 48.64		& 45.39	& 39.74		\\
& FuseGPT & 25\%	& 33.44	& 23.40	& 52.92	& 50.68		& 37.05	& 39.50		\\
\bottomrule
\end{tabular}
}
\end{table*}

\end{document}